\pdfoutput=1

\documentclass[11pt]{article}
\usepackage{marginnote}

\usepackage[preprint]{acl}

\usepackage{times}
\usepackage{latexsym}
\usepackage{tabularx}
\usepackage{tablefootnote}

\usepackage{enumitem}
\usepackage{listings}
\usepackage{balance}

\usepackage[T1]{fontenc}

\usepackage[utf8]{inputenc}

\usepackage{microtype}

\usepackage{inconsolata}

\usepackage{graphicx}
\usepackage{amsmath}

\usepackage{lipsum}
\usepackage{booktabs}

\usepackage{tikz}
\usepackage{fontawesome5}

\usepackage[colorinlistoftodos,textsize=tiny,disable]{todonotes}

\newcommand{\LightbulbOff}{%
  \begin{tikzpicture}[baseline=-.75ex]
    \node[inner sep=0pt] (L) {\faLightbulb[regular]};
    \draw[line width=1pt] (L.south west) -- (L.north east);
  \end{tikzpicture}%
}
\newcommand{\LightbulbOn}{%
  \begin{tikzpicture}[baseline=-.75ex]
    \node[inner sep=0pt] (L) {\faLightbulb[regular]};
  \end{tikzpicture}%
}
\newcommand{\thinking}{\LightbulbOn\hspace{1mm}}
\newcommand{\nothinking}{\LightbulbOff\hspace{1mm}}

\title{ClaimCheck: Real-Time Fact-Checking with Small Language Models}

\author{Akshith Reddy Putta\thanks{Equal contribution}\thanks{Author is a UTA affiliate attending Coppell High School.}, Jacob Devasier\footnotemark[1], Chengkai Li\thanks{Corresponding author} \\
  University of Texas at Arlington \\
  \texttt{\{akshith.putta, jacob.devasier, cli\}@uta.edu} \\}

\begin{document}
\maketitle 
\begin{abstract}
We introduce ClaimCheck, an LLM-guided automatic fact-checking system designed to verify real-world claims using live Web evidence and small language models. Unlike prior systems that rely on large, closed-source models and static knowledge stores, ClaimCheck employs a transparent, stepwise verification pipeline that mirrors human fact-checking workflows consisting of Web search query planning, Web-based evidence retrieval and summarization, evidence synthesis and re-retrieval, and claim verdict evaluation. Each module is optimized for small LLMs, allowing the system to deliver accurate and interpretable fact-checking with significantly lower computational requirements. Despite using a much smaller Qwen3-4B model, ClaimCheck achieves state-of-the-art accuracy of 76.4\% on the AVeriTeC dataset, outperforming previous approaches using LLaMA3.1 70B and GPT-4o. Extensive ablations demonstrate that careful modular design and prompting strategies can overcome the limitations of smaller LLMs. To promote accessibility and transparency, we provide a public demo at \url{https://idir.uta.edu/claimcheck}.
\end{abstract}

\section{Introduction}
The proliferation of misinformation across digital platforms has created an urgent need for accessible, reliable fact-checking tools. While manual fact-checking by professional organizations remains the gold standard, it is time-intensive and cannot scale to meet the volume of claims requiring verification. Recent automated fact-checking systems have achieved strong performance by combining retrieval-augmented generation with large language models (LLMs)~\cite{rothermel-etal-2024-infact, yoon-etal-2024-hero, malon-2024-multi}, but these approaches face significant barriers to widespread adoption: they typically depend on large, closed-source models that are computationally expensive, monetarily prohibitive, or challenging to deploy~\cite{schlichtkrull-etal-2024-automated, Braun2024DEFAMEDE}. 

In this work, we present ClaimCheck, an LLM-guided automatic fact-checking system that enables both experts and non-experts to verify real-world claims using real-time Web evidence and transparent, modular reasoning. Given a claim, the system displays the fact-checking process in real-time on a user-friendly interface and produces a step-by-step interpretable report that details how it planned searches, gathered and summarized evidence, synthesized information, and arrived at a final verdict.

\begin{figure*}[t]
  \centering
  \includegraphics[width=0.93\linewidth]{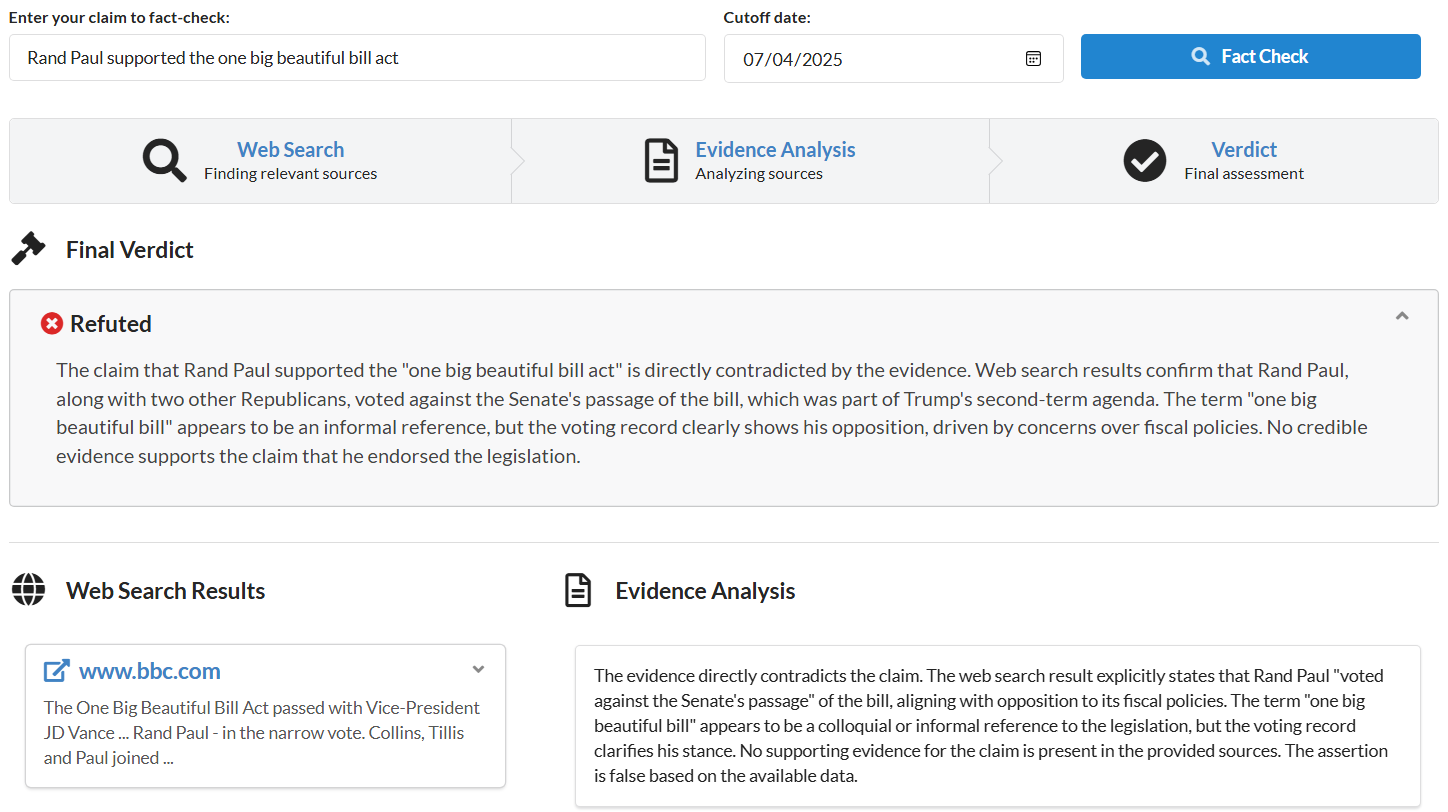}
  \caption{\label{fig:demo-full} An example of our demo on the claim ``Rand Paul supported the one big beautiful bill act.'' A list of evidence articles is shown in the bottom left, a synthesized analysis of the evidence is shown in the bottom right, and the final verdict produced by our system is shown in the middle. Each module is populated as they are completed, with a progress bar shown at the top.}
\end{figure*}

A central objective of ClaimCheck is to democratize access to trustworthy fact-checking tools by demonstrating that effective automated fact-checking can be achieved with substantially smaller, more accessible models than previous systems. Smaller LLMs offer compelling advantages including reduced computational requirements, lower operational costs, and suitability for deployment on local or edge devices. However, such models commonly exhibit lower task-specific performance and limited reasoning capabilities compared to their larger counterparts~\cite{putta-etal-2025-claimcheck}. Previous efforts using smaller LLMs for fact-checking have encountered notable challenges. For instance, \citet{putta-etal-2025-claimcheck} using the Qwen2.5 7B model struggled with effectively synthesizing evidence and predicting the verdict for fact verification. Building on the modular approach established by DEFAME~\cite{Braun2024DEFAMEDE}, this work demonstrates that these limitations can be systematically addressed through specialized modular design optimized for smaller LLMs that decomposes fact-checking into manageable components. 

ClaimCheck mirrors human fact-checking workflows through five core stages: query planning, evidence retrieval, evidence summarization/filtering, evidence synthesis and re-retrieval, and claim evaluation. These modules communicate via a centralized fact-checking report that serves as both an internal coordination mechanism and a user-facing explanation of the reasoning process. This design enables users to inspect and understand the system's decision-making at each step, addressing critical transparency requirements for fact-checking. 

We evaluate ClaimCheck on the AVeriTeC dataset~\cite{schlichtkrull-etal-2024-automated}, applying strict temporal cutoffs to prevent data leakage and reflect realistic verification conditions. Despite using a much smaller model (Qwen3-4B), ClaimCheck achieves state-of-the-art accuracy (76.4\%), outperforming prior approaches---PASS-FC~(72.0\%)~\cite{zhuang2025passfc} and HerO~(75.2\%)~\cite{yoon-etal-2024-hero}---which rely on substantially larger models (e.g., LLaMA3.1 70B and GPT-4o) and use pre-fetched knowledge stores. Through comprehensive ablation studies, we demonstrate that stepwise reasoning within synthesis and evaluation modules is crucial for performance, and that careful module design and prompting strategies can overcome the limitations associated with smaller LLMs. 

A public demo for ClaimCheck is provided at \url{https://idir.uta.edu/claimcheck} and we provide an example of its outputs in Figure~\ref{fig:demo-full}. We also open-source our code at \url{https://github.com/idirlab/claimcheck}.  \todo[color=yellow]{And I suppose you are aware the demo URL is currently not working.}

\begin{figure}[t]
  \centering
  \includegraphics[width=0.85\linewidth]{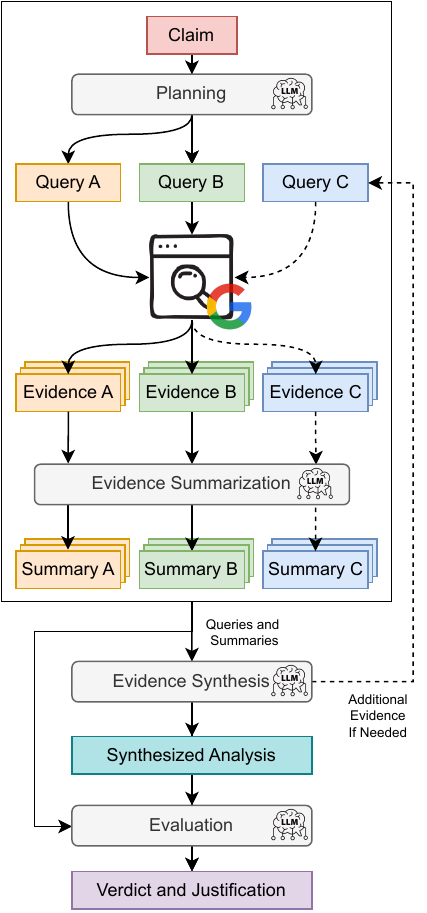}
  
  \caption{\label{fig:claimcheck-system}Overview of ClaimCheck's LLM-guided pipeline. The dashed arrow indicates the system's ability to perform search queries to retrieve necessary additional evidence to fact-check the claim.}
  \vspace{-8mm}
\end{figure}

\section{ClaimCheck}
ClaimCheck is a LLM-guided automatic fact-checking system. It employs a modular architecture where each component performs a distinct function in the verification workflow (Figure~\ref{fig:claimcheck-system}). We utilize Qwen3-4B~\cite{yang2025qwen3technicalreport} with thinking enabled across all modules as our LLM to ensure comprehensive reasoning at each step. The LLM prompts used for each module are presented in Appendix~\ref{app:prompts} and are largely based on DEFAME's~\cite{Braun2024DEFAMEDE} prompts. 

\subsection{System Architecture}
The ClaimCheck system operates through a structured pipeline mirroring human fact-checking practices while leveraging the scalability of automated systems.\todo{In Figure 2, the word ``Execution'' is not mentioned. Unnecessary to use a magnifier icon; you are not using such icons in other places. Even the Google icon might not be necessary.}\todo{Instead of ``execution'', would it be more clear to call it ``search query execution''} The approach begins by analyzing the claim and constructing Web search queries to gather evidence (\textit{planning}). Next, the system uses the queries to find and collect relevant articles (\textit{execution}) and summarizes key points about the claim from the articles (\textit{evidence summarization}). Then, all of the evidence is synthesized to a single cohesive analysis to determine if additional information is needed (\textit{evidence synthesis}) and collect it if so (\textit{execution}). If no further evidence is needed, the system proceeds to fact verification. Finally, the system uses the synthesized evidence to determine the veracity of the claim and provide a justification accordingly (\textit{evaluation}). 

\paragraph{Fact-Checking Report.}
ClaimCheck generates a comprehensive fact-checking report for each input claim. This report serves as a central artifact that facilitates communication between modules by recording their intermediate outputs in a structured format. This enables later modules to reference earlier outputs and ensures consistent, interpretable outputs for users. Once the fact-checking process is complete, users can download the report to review the system’s internal reasoning in each module.  \todo[color=yellow]{The report itself is not mentioned in the modules or architecture diagram. Is there a way to make it more clear how it fits in the whole architecture? Currently this short paragraph is disconnected from both the content before and after it.}

\subsection{Modules} 
\paragraph{Planning.} The \textit{planning} module utilizes the LLM to understand the input claim and construct a set of Web search queries to collect evidence to fact-check the claim. The LLM is instructed~(Listing~\ref{lst:plan_prompt} in Appendix~\ref{app:prompts}) to consider different aspects of the claim in case a claim may have multiple sub-claims. While we do not explicitly instruct the LLM to perform claim decomposition, as is common in many previous works~\cite{Braun2024DEFAMEDE, guo-etal-2022-survey, iqbal-etal-2024-openfactcheck}, Qwen3 often breaks down the claim in its thinking. 

\begin{figure}
  \centering
  \includegraphics[width=0.75\linewidth]{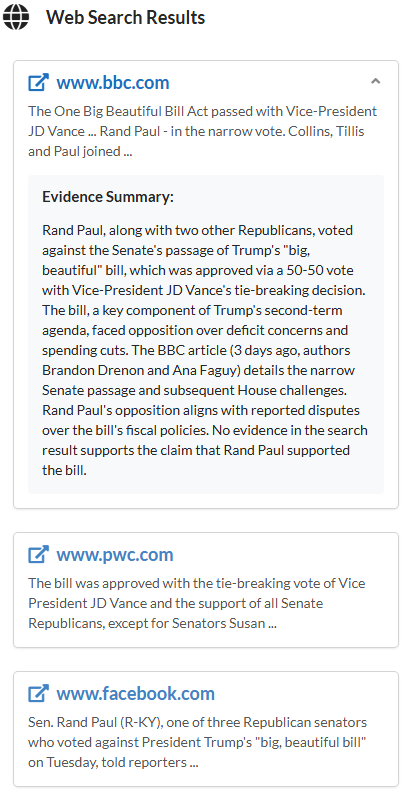}
  \caption{\label{fig:demo-search}An example of a summarized article collected using ClaimCheck's Web evidence search for the claim ``Joe Biden voted for the Iraq War.''}
\end{figure}

\todo{Figure~\ref{fig:demo-search} cpations says ``a summarized article'', but now it has 3 articles.}

\paragraph{Execution.} The \textit{execution} module carries out the Web search using the queries created in the \textit{planning} module. We utilize Serper~\footnote{\url{https://serper.dev/}} for executing our queries using the Google search engine, returning URLs and snippets for each article. In our demo, we limit this to only the top-3 results to significantly increase the processing speed of the system. We present this list of URLs and snippets in the Web search results (Figure~\ref{fig:demo-search}). 

\paragraph{Evidence Summarization.} The \textit{evidence summarization} module uses the LLM~(Listing~\ref{lst:summ_prompt}) to individually extract and summarize the relevant information from each collected article and discard articles which are not helpful for fact-checking the claim. Figure~\ref{fig:demo-search} shows how the evidence summary for each collected article is displayed to the user. 

\paragraph{Evidence Synthesis.} The \textit{evidence synthesis} module uses the LLM~(Listing~\ref{lst:synth_prompt}) synthesizes all of the summarized evidence into a single coherent analysis that consolidates supporting and contradicting information, identifies patterns across sources, highlights key factual information with their supporting evidence, and assesses the overall reliability and consistency of the collected information. After the analysis is produced, the LLM will determine whether additional Web search is needed, taking into account any gaps in evidence and previous queries, and produce a Web search query if so. This will trigger the system to rerun the \textit{execution} and \textit{evidence summarization} modules. 

\paragraph{Evaluation.} The \textit{evaluation} module uses the LLM~(Listing~\ref{lst:eval_prompt}) to evaluate the output of the synthesized evidence and assigns a final verdict to the claim along with an explanation of the predicted veracity. Figure~\ref{fig:demo-full} shows an example of ClaimCheck's predicted verdict and produced explanation for the claim ``Joe Biden voted for the Iraq War.'' 

\section{Experiments}

\subsection{Dataset}
AVeriTeC~\cite{Schlichtkrull2023AVeriTeCAD} is a benchmark dataset for real-world automatic fact-checking. This dataset contains 4,568 authentic claims sourced from 50 different fact-checking organizations. Each claim is annotated with (1) one of four verdicts: Supported, Refuted, Conflicting Evidence/Cherrypicking, or Not Enough Evidence, (2) Question-answer pairs grounded in Web evidence which justify the verdict on the claim, (3) textual justifications which summarize how the collected evidence supports the verdict, and (4) metadata about the speaker, publisher, publication date, and location of the claim.\todo[color=cyan]{publisher is where a claim was first stated.} These annotations are separated into Train (3,068 samples), Development (500 samples), and Test (1000 samples) partitions. The authors also released a knowledge store containing roughly 1,000 Web articles for each claim to support offline retrieval systems.


\begin{table}[t]
\centering
\resizebox{\linewidth}{!}{
\begin{tabular}{lcc}
    \toprule
    \textbf{Framework} & \textbf{Accuracy} & \textbf{LLM}\\ 
    \midrule
    \faBan~GPT-4o mini & 46.8\% & GPT-4o mini\\ 
    \faBan~Qwen3 4B & 52.0\% & Qwen3-4B\\ 
    \faBan~GPT-4o & 53.2\% & GPT-4o\\
    \midrule
    \faDatabase~InFact & 72.4\% & GPT-4o\\ 
    \faDatabase~HerO & \underline{75.2\%} & Llama3.1 70B\\ 
    \midrule
    \faGlobe~Papelo & 41.5\% & GPT-4o mini\\
    \faGlobe~ClaimCheck* & 67.0\% & OpenAI o4-mini\\ 
    \faGlobe~GPT-4o mini & 70.4\% & GPT-4o mini\\
    \faGlobe~DEFAME & 70.5\% & GPT-4o\\
    \faGlobe~PASS-FC & 72.0\% & GPT-4o\\
    \faGlobe~GPT-4o & 74.6\% & GPT-4o\\
    \faGlobe~ClaimCheck* & 76.0\% & Qwen3-32B\\ 
    \faGlobe~ClaimCheck & \textbf{76.4\%} & Qwen3-4B\\ 
    \bottomrule
    \multicolumn{3}{l}{\small * Results only evaluated on a subset of 100 claims.}
\end{tabular}
}
\caption{Verdict prediction accuracy of different frameworks on the AVeriTeC development dataset. Knowledge sources used are indicated by their icons. \faGlobe~indicates using Web search for evidence, \faDatabase~indicates using the knowledge store, and \faBan~indicates using no evidence.}
\label{tab:framework_accuracy}

\vspace{-2mm}
\end{table}

\subsection{Experimental Setup}
\label{sec:experimental-setup}
We evaluate ClaimCheck on the development split of the AVeriTeC dataset, as the test set lacks gold verdict labels. The development set has a similar class distribution to other sets: 24.4\% \textit{Supported}, 61.0\% \textit{Refuted}, 7.6\% \textit{Conflicting Evidence/Cherrypicking}, and 7.0\% \textit{Not Enough Evidence}. To prevent temporal data leakage during Web retrieval, we restrict our search to documents published before the claim’s annotated publication date. This ensures the system does not access information that would not have been available at the time the claim was made, such as later news coverage, follow-up investigations, or even the professional fact-check that produced the verdict label itself.

\todo{Is this the place where you explain cutoff date? You are not relating it to the cutoff option on the user interface.}

We report veracity prediction accuracy, defined as the proportion of claims for which the system correctly predicts the verdict. Our system is compared against the top-performing submissions from the 2024 AVeriTeC shared task~\cite{schlichtkrull-etal-2024-automated}: {InFact}~\cite{rothermel-etal-2024-infact}, {Papelo}~\cite{malon-2024-multi}, and {HerO}~\cite{yoon-etal-2024-hero}, as well as two recent fact-checking systems: {DEFAME}~\cite{Braun2024DEFAMEDE} and {PASS-FC}~\cite{zhuang2025passfc}. We group these systems based on their evidence retrieval method (Web search vs. knowledge store) to ensure a fair comparison (see Table~\ref{tab:framework_accuracy}). 

\paragraph{Compared Baselines.} Among these, {HerO}---the previous best-performing system---generates hypothetical fact-checking documents from a given claim and retrieves semantically similar evidence from the AVeriTeC knowledge store for fact verification using LLaMA 3.1 70B. {InFact} decomposes each claim into up to 10 sub-questions and queries the knowledge store to answer them, then the answers are synthesized by an LLM to predict a final verdict. {Papelo} iteratively prompts an LLM to generate and answer questions using Web search results to support claim verification.~\footnote{{Papelo}'s final system was instructed not to predict the \textit{Not Enough Evidence} or \textit{Conflicting Evidence/Cherrypicking} classes. For a fair comparison, we report results from their best-performing 4-class model, which has lower accuracy.}

{PASS-FC}, the previous best-performing Web evidence-based system, reformulates a given claim into \textit{comprehensive claims}---adding temporal and entity-specific context---and then uses an iterative self-reflection module to determine whether further evidence is required. {DEFAME} employs a multimodal LLM (GPT-4o) to orchestrate iterative planning, retrieval, and analysis for fact-checking text and image-based claims and was a significant source of inspiration for this work.\todo{Section 2 says our prompts are based on those from DEFAME. But where are we explaining in what ways ClaimCheck is different and makes new contributions?}

We also include additional baselines using GPT-4o and GPT-4o mini with OpenAI's Web search tool (see Appendix~\ref{app:reproducibility}). While these models cannot be constrained by a cutoff date---making them susceptible to significant data leakage---we include them for completeness and because no previous work has explored their usage. Note that, to the best of our knowledge, a Qwen3 Web search equivalent model does not exist. Finally, to assess the impact of internal LLM knowledge without external evidence, we also report the zero-evidence performance of GPT-4o, GPT-4o mini, and Qwen3-4B (taken from \citealp{putta-etal-2025-claimcheck}).

\subsection{Model Performance Comparison}
Table~\ref{tab:framework_accuracy} presents the verdict prediction accuracy of ClaimCheck and other systems, grouped by evidence source. ClaimCheck achieves state-of-the-art performance with 76.4\% accuracy using Qwen3-4B, outperforming all prior systems including those using significantly larger models. We also include results of subsequent experiments from Section~\ref{sec:model-scale}. These experiments use a random subset of 100 claims from AVeriTeC's development dataset due to monetary and time constraints.

\paragraph{Comparison with Web-based Systems.} 
ClaimCheck substantially outperforms existing Web-based fact-checking systems, achieving 76.4\% accuracy compared to PASS-FC (72.0\%), DEFAME (70.5\%), and GPT-4o with Web search (74.6\%). Notably, GPT-4o's performance benefits from temporal data leakage, as it cannot be constrained by publication cutoff dates and often retrieves content from annotated fact-check articles in AVeriTeC.  In contrast, ClaimCheck enforces strict temporal constraints, filtering out any evidence published after each claim's annotated publication date.

\paragraph{Comparison with Knowledge Store Systems.}
Despite relying on live Web retrieval rather than curated knowledge stores, ClaimCheck outperforms leading knowledge store-based systems, including HerO (75.2\%) and InFact (72.4\%). These systems depend on substantially larger models—LLaMA3.1 70B and GPT-4o, respectively—and benefit from access to gold evidence drawn directly from the AVeriTeC knowledge store. In contrast, ClaimCheck operates under more realistic constraints while achieving higher accuracy (76.4\%).

\paragraph{Zero-Evidence Baselines.}
To assess the impact of internalized knowledge in large language models, we evaluate several models in a zero-evidence setting. GPT-4o achieves the highest accuracy among them (53.2\%), followed by Qwen3-4B (52.0\%) and GPT-4o mini (46.8\%). While these scores are substantially lower than those of evidence-augmented systems, they suggest that modern LLMs encode a nontrivial amount of veracity-relevant world knowledge. Notably, Qwen3-4B's performance is especially strong given its relatively small size, further motivating its use within the {ClaimCheck} system.


\begin{table}
    \centering
    \resizebox{\linewidth}{!}{
        \begin{tabularx}{\linewidth}{l c}
            \toprule
            \textbf{Configuration} & \textbf{Accuracy} \\
            \midrule
            \textbf{All modules Think (baseline)} & 75.0\% \\
            \nothinking Planning & 71.0\% \\
            \nothinking Evidence Summarization & 66.0\% \\
            \nothinking Evidence Synthesis & 72.0\% \\
            \nothinking Evaluation & 65.0\% \\
            \midrule
            \textbf{All modules No-Think (baseline)} & 54.0\% \\
            \thinking Planning & 59.0\% \\
            \thinking Evidence Summarization & 60.0\% \\
            \thinking Evidence Synthesis & 60.0\% \\
            \thinking Evaluation  & 61.0\% \\
            \bottomrule
        \end{tabularx}
    }
    \caption{Ablation study of reasoning module configurations on the random subset of 100 claims from AVeriTeC's development set. Each row modifies the \textit{think} setting of a single module while others remain fixed. \protect\thinking indicates that thinking is enabled only for a particular module, and vice versa with \protect\nothinking.}
    \label{tab:ablation_study}

    \vspace{-2mm}
\end{table}




\subsection{Hybrid Thinking Ablation Study}
\label{sec:ablation}
A key innovation\todo{This is the first time this ``key innovation'' is mentioned. And it was not explained earlier or now. Shouldn't this be explained earlier?} in ClaimCheck is the use of Qwen3's hybrid thinking capability. We conduct an ablation study on the previously-mentioned 100 randomly sampled claims from the AVeriTeC development set to understand the contribution of thinking in each module. Table~\ref{tab:ablation_study} summarizes the results of this experiment.

The fully thinking configuration achieves 75.0\% accuracy, substantially outperforming the all no-think baseline (54\% accuracy). This 21 percentage point improvement demonstrates the significant value of reasoning capabilities in fact-checking tasks. When disabling thinking in individual modules from the all-think baseline, we observe varying performance impacts. The \emph{evidence summarization} and \emph{evaluation} modules show the largest degradation (9--10 percentage points), indicating these components are most critical for effective reasoning. The \emph{planning} and \emph{evidence synthesis} modules show smaller reductions (3--4 percentage points), suggesting these tasks are less reasoning-intensive.

Starting from the all no-think baseline, enabling thinking in any single module provides consistent but modest improvements (5--7 percentage points), with no single module dominating. This suggests that while each module benefits from thinking, the full pipeline requires comprehensive reasoning for optimal performance. Based on these findings, we enable thinking across all modules in the final system, as the performance benefits outweigh the minimal computational overhead (Table~\ref{tab:model_compute_performance}).

\subsection{Model Scale Analysis}
\label{sec:model-scale}
We investigate whether larger models improve performance by comparing the Qwen3-4B and Qwen3-32B variants of ClaimCheck. Surprisingly, we observe no significant performance difference between the two model sizes (76.4\% vs. 76.0\% accuracy, as shown in Table~\ref{tab:framework_accuracy}), suggesting that model scale alone is not the primary driver of performance in our pipeline. This finding supports our hypothesis that effective fact-checking systems benefit more from high-quality evidence retrieval and careful architectural design than from raw model capacity. However, when using OpenAI's o4-mini as the base LLM, accuracy drops significantly to 67.0\%. This degradation likely stems from our prompting strategy being optimized for Qwen3 models, highlighting the importance of model-specific optimization in modular architectures.


\begin{table}[t]
    \centering
    \resizebox{\linewidth}{!}{
        \begin{tabular}{lll}
            \toprule
            \textbf{Model} & \textbf{Compute} & \textbf{Avg. Time} \\ 
            \midrule
            \thinking Qwen3-32B  & H100 & 127 seconds \\ 
            \thinking Qwen3-4B  & RTX 4070 & 79 seconds \\ 
            \thinking Qwen3-4B & H100 & 66 seconds \\ 
            \nothinking Qwen3-4B  & H100 & 61 seconds \\ 
            \thinking OpenAI o4-mini  & OpenAI API & 32 seconds \\ 
            \midrule
            \nothinking GPT-4o + Search  & OpenAI API & 1 second \\ 
            \nothinking GPT-4o-mini + Search & OpenAI API & {0.75} seconds \\ 
            \bottomrule
        \end{tabular}
    }
\caption{Throughput of ClaimCheck using different LLMs. \protect\thinking indicates that thinking is used for a particular model, and vice versa with \protect\nothinking.}
\label{tab:model_compute_performance}
\vspace{-2mm}
\end{table}

\paragraph{Computational Efficiency Analysis.}
Table~\ref{tab:model_compute_performance} presents the computational efficiency of different ClaimCheck configurations. ClaimCheck with Qwen3-4B on average processes each claim in 66 seconds on an H100 GPU and 79 seconds on a more accessible RTX 4070. Using Qwen3-32B doubles the runtime while providing no meaningful accuracy improvement. The thinking mechanism adds minimal overhead, requiring only 5 more seconds compared to the non-thinking model. 

GPT-4o and GPT-4o-mini with Web search achieve extremely fast processing times due to large-scale retrieval infrastructure and high-speed inference resources. However, their use of temporally leaked evidence, as discussed in Section~\ref{sec:experimental-setup}, compromises their performance metrics. 

\section{Conclusion} 

This work presents ClaimCheck, an LLM-guided fact-checking system that verifies real-world claims using live Web evidence and small, accessible models. By decomposing fact-checking into discrete, interpretable stages, ClaimCheck enables effective reasoning with the Qwen3-4B model, outperforming systems powered by much larger LLMs on the AVeriTeC dataset. Our ablation studies highlight the critical role of reasoning, particularly in the evidence synthesis and evaluation modules, which contribute significantly to the system’s effectiveness. Our results show that thoughtful architectural design and specialized prompting can overcome limitations typically associated with smaller LLMs.

ClaimCheck offers a promising direction for democratizing access to trustworthy, transparent fact verification. With a public demo and downloadable fact-checking reports, the system promotes transparency and empowers both experts and non-experts to assess the truthfulness of claims.


\section*{Ethics and Risks} 
\paragraph{Misinformation Amplification.} 
While ClaimCheck is designed to detect and mitigate misinformation, its reliance on live Web search introduces potential risks. Specifically, the system may inadvertently retrieve and summarize low-quality or misleading sources, especially if those are ranked highly by search engines. Although our modular pipeline includes filtering and synthesis stages to reduce this risk, the system cannot guarantee that all retrieved content is accurate or representative. To mitigate this, we provide transparent reporting so users can trace each verdict back to its evidence.

\paragraph{Over-reliance and Misuse.}
Users may overestimate the capabilities or correctness of the system, especially given its step-by-step justifications and high reported accuracy. However, the system is not infallible and should not be used as a substitute for human judgment in high-stakes domains such as legal decisions, medical claims, or policy debates. To discourage misuse, we explicitly label the system as experimental and provide downloadable reports to encourage further human review.

\paragraph{Bias and Fairness.}
The LLMs powering ClaimCheck are pretrained on Web-scale data and may reflect underlying social, political, or cultural biases. These biases can surface in query planning, evidence summarization, and especially in the final evaluation stage. Although we use a relatively small and accessible model (Qwen3-4B), its reasoning remains influenced by the distributions in its training data. We partially address this through modular decomposition and prompt engineering, but biases in evidence selection or synthesis remain a concern---particularly for politically sensitive or underrepresented topics.


\balance

\bibliography{anthology,custom}

\appendix \vspace{2mm}

\section{Reproducibility}
\label{app:reproducibility}
\paragraph{LLM Hyperparameters.} ClaimCheck uses LLMs without finetuning for any tasks, with temperature set to 0.6 and top-\textit{p} to 0.95 for all Qwen3 models, and temperature set to 0.3 and top-\textit{p} to 1.0 for o4-mini. We used Ollama (\url{https://ollama.com/}) to run Qwen3 models and the OpenAI API to run o4-mini. For the Web search-enabled GPT-4o/GPT-4o mini, we used \texttt{gpt-4o-search-preview-2025-03-11} and \texttt{gpt-4o-mini-search-preview-2025-03-11}.

\paragraph{Ablation random sample.} The same random sample is used across all ablations with a fixed random seed of 42 for reproducibility. 

\paragraph{LLM Model Costs.} Approximate inference cost per 100 claims is \$3 for GPT-4o with Web search (due to both API and search engine calls) and \$1 for ClaimCheck when using Fireworks AI as a serverless inference provider. Table~\ref{tab:model_compute_performance} excludes Qwen3 with Web search, as we did not run those experiments. We focus on GPT-based models for this comparison due to their native support for integrated Web search, which Qwen3 currently lacks.

\section{Error Analysis}

A key challenge observed in our system involves the behavior of the Qwen3-4B model during the \textit{evidence summarization} stage. Despite explicit instructions to filter and concisely summarize only the most relevant information from each article, the model often produces overly long summaries that reiterate large portions of the input text. Our qualitative analysis suggests that when the prompt becomes too long—typically due to the accumulation of article content and prior module outputs—Qwen3-4B tends to prioritize summarizing the entire prompt over following the specified summarization task. This behavior introduces noise into the downstream synthesis and evaluation stages and occasionally causes prompt overflows, especially during iterative re-retrieval.

In addition to model-specific issues, limitations in the AVeriTeC dataset itself may impose an upper bound on achievable accuracy—likely capping system performance near 80\%.\todo{Unclear how we derive this estimate of 80\%.} Many claims hinge on nuanced or ambiguous interpretations that challenge even human annotators, as there are not exact guidelines reported in AVeriTeC to describe the verdicts. For example, consider the following pair of claims:

\begin{itemize}
    \item \textbf{Claim 1} (October 27, 2020): \textit{"Joe Biden wants to ban fracking"} — labeled \textbf{Refuted}.
    \item \textbf{Claim 2} (October 23, 2020): \textit{"Joe Biden said he wants to ban fracking in the US"} — labeled \textbf{Conflicting Evidence/Cherrypicking}.
\end{itemize}

Both claims reference similar content and timeframes, yet are assigned different verdicts. The justification for the “Refuted” label is that Biden never explicitly stated an intent to ban fracking, whereas the “Conflicting” label is justified by presenting both a 2019 quote suggesting eventual elimination of fossil fuels and later statements ruling out a fracking ban. Such complexities make it difficult for any automated system—even with perfect evidence—to confidently assign the correct verdict. As a result, future benchmarks may need clearer annotation protocols and better handling of ambiguous claims to further push fact-checking performance beyond current ceilings.

\section{Prompts}
\label{app:prompts}
We adopt the prompting structure from DEFAME~\cite{Braun2024DEFAMEDE} and adapt it to better suit smaller models such as Qwen3-4B. Specifically, we simplify instructions, reduce verbosity, and structure prompts to minimize context window overflow, which smaller LLMs are more susceptible to. 

The prompts are depicted in Listings~\ref{lst:plan_prompt}--~\ref{lst:eval_prompt}. Particularly, the following verdict descriptions are embedded in the ``Rules'' section of the evaluation prompt (Listing~\ref{lst:eval_prompt}).

\vspace{2mm}
\ttfamily
\small

Supported
- The claim is directly and clearly backed by strong, credible evidence. Minor uncertainty or lack of detail does not disqualify a claim from being Supported if the main point is well-evidenced.
- Use Supported if the overall weight of evidence points to the claim being true, even if there are minor caveats or not every detail is confirmed.

\vspace{2mm}

Refuted
- The claim is contradicted by strong, credible evidence, or is shown to be fabricated, deceptive, or false in its main point.
- Use Refuted if the central elements of the claim are disproven, even if some minor details are unclear.
- Lack of any credible sources supporting the claim does not mean ``Not Enough Evidence''---it means the claim is Refuted.

\vspace{2mm}

Conflicting Evidence/Cherrypicking
- Only use this if there are reputable sources that directly and irreconcilably contradict each other about the main point of the claim, and no clear resolution is possible after careful analysis.
- Do NOT use this for minor disagreements, incomplete evidence, or if most evidence points one way but a few sources disagree.

\vspace{2mm}

Not Enough Evidence
- Only use this if there is genuinely no relevant evidence available after a thorough search, or if the claim is too vague or ambiguous to evaluate.
- Do NOT use this if there is some evidence, even if it is weak, or if the claim is mostly clear but not every detail is confirmed.
- This is a last-resort option only.

\normalsize
\rmfamily

\todo{Include some vertical space between Listing cpation and the box} 

\lstset{
    basicstyle=\ttfamily\small,
    backgroundcolor=\color{gray!10}, 
    linewidth=\linewidth,
    breaklines=true,
    frame=single, 
    rulecolor=\color{black}, 
    showstringspaces=true, 
    numbers=left, 
    numberstyle=\tiny\color{gray}, 
    xleftmargin=0em, 
    numbers=none, 
    columns=fullflexible,
    breakindent=0pt,
    moredelim=**[is][\color{gray!70}]{@}{@} 
}
\begin{figure*}[th]

\begin{lstlisting}[caption={Prompt for Planning}, label=lst:plan_prompt]
Instructions
The available knowledge is insufficient to assess the Claim. Therefore, propose a set of actions to retrieve new and helpful evidence. Adhere to the following rules:

- The actions available are listed under Valid Actions, including a short description for each action. No other actions are possible at this moment.
- For each action, use the formatting as specified in Valid Actions.
- Include all actions in a single Markdown code block at the end of your answer.
- Propose as few actions as possible but as much as needed. Do not propose similar or previously used actions.
- Consider Both Modalities Equally: Avoid focusing too much on one modality at the expense of the other, but always check whether the text claim is true or false.
- Compare Image and Caption: Verify the context of the image and caption.

Valid Actions:
web search: Run an open web search for related webpages.

Examples:
web_search("New Zealand Food Bill 2020")

Record:
{record}

Claim: {claim}
Your Actions:

\end{lstlisting}
\end{figure*}

\begin{figure*}

\begin{lstlisting}[caption={Prompt for Evidence Summarization}, label=lst:summ_prompt]
Instructions
In order to find evidence that helps your fact-check, you just ran a web search, which yielded a Search Result. Your task right now is to summarize the Search Result concisely in at most 5 sentences, only including information that is relevant to the Claim you are checking.

What to include:
- Information that might be useful for fact-checking the claim (see Record).
- If available: the release date as well as the author or the publisher (e.g., the media company) of the search result.

Do NOT include:
- Advertisements.
- Any other information unrelated to the Record or the Claim.

Additional Rules:
- Do not add any additional information besides the information in the Search Result. Also, do not add any information that is not related to the claim, even if it is mentioned in the Search Result.
- If the Search Result doesn't contain any relevant information for the fact-checking work, print only one word in capital letters, do not include anything else: NONE.
- Keep your writing style consistent with the provided Examples.
- Try to filter out relevant information even if the Search Result is in a different language.

Claim: {claim}

Evidence:
{url}
{search_result}

Record:
{record}


Your Summary:
\end{lstlisting}
\end{figure*}

\begin{figure*}[th]

\begin{lstlisting}[caption={Prompt for Evidence Synthesis}, label=lst:synth_prompt]
Instructions
You just retrieved new Evidence. Now, analyze the Claim's veracity using the evidence. Always adhere to the following rules:
- Focus on developing new insights. Do not repeat larger parts from the Record. Do not restate the Claim.
- Write down your thoughts step-by-step. Whenever necessary, you may elaborate in more detail.
- Depending on the topic's complexity, invest one to three paragraphs. The fewer, the better.
- If you find that there is insufficient information to verify the Claim, explicitly state what information is missing.
- If you cite web sources, always refer to them by including their URL as a Markdown hyperlink.
- Use information only from the recorded evidence:
- Avoid inserting information that is not implied by the evidence. You may use commonsense knowledge, though.

If it is extremely necessary to retrieve more evidence, you can propose actions to the user. If not necessary, do not add anything else other than the reasoning. 

Adhere to the following rules:
- The actions available are listed under Valid Actions, including a short description for each action. No other actions are possible at this moment.
- For each action, use the formatting as specified in Valid Actions.
- Propose as few actions as possible but as much as needed. Do not propose similar or previously used actions.
- Include all actions in a single Markdown code block at the end of your answer.

Valid Actions:
web search: Run an open web search for related webpages.

Examples:
web_search("New Zealand Food Bill 2020")

Record:
{record}

Your Analysis:
\end{lstlisting}
\end{figure*}

\begin{figure*}[th]

\begin{lstlisting}[caption={Prompt for Evaluation}, label=lst:eval_prompt]
Instructions
Determine the Claim's veracity by following these steps:
1. Briefly summarize the key insights from the fact-check (see Record) in at most one paragraph.
2. Write one paragraph about which one of the Decision Options applies best. Include the most appropriate decision option at the end and enclose it in backticks
like `this`.

Decision Options:
Supported|Refuted|Conflicting Evidence/Cherrypicking|Not Enough Evidence

Rules:
{verdict_descriptions}

Record:
{record}
Your Judgement:
\end{lstlisting}
\end{figure*}
\end{document}